# Platoon trajectories generation: A unidirectional interconnected LSTM-based car-following model

Yangxin Lin, Ping Wang, *Senior Member, IEEE,* Yang Zhou, Fan Ding,
Chen Wang, Huachun Tan, *Member, IEEE*

*Abstract*—Car-following models have been widely applied and made remarkable achievements in traffic engineering. However, the traffic micro-simulation accuracy of car-following models in a platoon level, especially during traffic oscillations, still needs to be enhanced. Rather than using traditional individual car-following models, we proposed a new trajectory generation approach to generate platoon level trajectories given the first leading vehicle's trajectory. In this paper, we discussed the temporal and spatial error propagation issue for the traditional approach by a car following block diagram representation. Based on the analysis, we pointed out that error comes from the training method and the model structure. In order to fix that, we adopt two improvements on the basis of the traditional LSTM-based car-following model. We utilized a scheduled sampling technique during the training process to solve the error propagation in the temporal dimension. Furthermore, we developed a unidirectional interconnected LSTM model structure to extract trajectories features from the perspective of the platoon. As indicated by the systematic empirical experiments, the proposed novel structure could efficiently reduce the temporal-spatial error propagation. Compared with the traditional LSTM-based car-following model, the proposed model has almost 40% less error. The findings will benefit the design and analysis of micro-simulation for platoon-level car-following models.

## I. INTRODUCTION

Micro-traffic simulation platform relies on car-following models to simulate the behavior of Human-Driven Vehicles (HDVs) and is widely used in most traffic applications, such as traffic operation [1] and control [2]–[4]. The performance of those applications on the micro-simulation platform is determined by the accuracy of the car-following models embedded in simulation to reproduce the actual driving behavior.

Car following models have been extensively studied and analyzed, which can be largely put into two categories, model-based parametric car-following model and model-free nonparametric car-following models.

Based on the prior knowledge of driving behavior, model-based parametric car-following models make specific assumptions about driving behavior. For example, Bando et al. proposed the Optimal Velocity (OV) model to formulate optimal following velocity in order to simulate the stop-and-go traffic [5]; Treiber et al. proposed Intelligent Driver Model (IDM), which decides the control strategy by considering both desired acceleration and spacing-related deceleration [6]; Laval and Leclercq reproduced traffic oscillation by involving various driving behavior (e.g. aggressive or timid driver behavior) [7]; Chen et al. develop a dynamic driving behavior to explain the mechanisms that induce traffic oscillations [8]. Though model-based parametric model has a great insight to interpret the traffic phenomenon (e.g. traffic oscillation [8], capacity drop [9], traffic hysteresis [10]) via the predefined car-following structure and parameters considering the drivers' characteristics in lane-changing [11]–[13], time-varying car-following response time [7], [8] etc., those model-based methods cannot fully capture the nuances of such a complex phenomenon. Nevertheless, due to the great physical explanative capability and less computational consumption, model-based car-following models are widely used in traffic simulation platforms, such as VISSIM, AIMSUN, and PARAMICS [14].

In contrast to the model-based models, model-free nonparametric car-following models, such as neural network (NN) based model [4], [15]–[19], could fit better vehicle-level driving behavior with reality, especially in terms of reproducing the trajectories. In general, NN-based models perform better than model-based models due to the flexible model structure and fit advances. They can automatically learn driving behavior from the field data without any artificial parameters [17]. To the authors' knowledge, the state of art NN-based car-following model is based on Recurrent Neural Network (RNN), which is a representative neural network for processing sequential data. By involving the mechanism of information passing among decision steps, RNN-based car-following models consider the memory effect or the prediction capability of drivers [15]. Researches have indicated that RNN-based car-following models have better performance on duplicating driving behavior than model-based models [15]–[17], [20]. Morton et

This work is financed by the National Key R&D Program in China (Grant No. 2018YFB1600600), Science and Technology Major Project, Transportation of Jiangsu Province, and the National Science Foundation of China (No. 61620106002).
(Corresponding author: Yang Zhou.)
Y. Lin is with the School of Software & Microelectronics, Peking University, Beijing 100871, China.
Y. Zhou is with the Department of Civil and Environmental Engineering, University of Madison, WI, USA. (email: zhou295@wisc.edu)
P. Wang is with National Engineering Research Center for Software Engineering, Beijing, China.
F. Ding is with the School of Transportation, Southeast University, Nanjing, China.
C. Wang is with Intelligent Transportation Research Center, Southeast University, Nanjing, China.
Huachun Tan is with the School of Transportation, Southeast University, Nanjing, China.

al. utilize LSTM network, which is known as the variant of RNN, to learn probability model of driving behavior [16]; Huang et al. develop car-following model by LSTM network, which shows better performance on reduplicating the asymmetric driving behavior than IDM [19]; Wang et al. design a multi-layer Gated Recurrent Unit (GRU) [21] network, which is a simplified version of LSTM, to capture the car-following behavior [15]. The strong learning capability gives NN-based models an opportunity to find potential mode and law in the empirical driving behavior data.

However, the accuracy issues of the aforementioned car-following models in a platoon level, especially under traffic oscillations, still needs to be enhanced as indicated by [8], [15], [17]. Misestimating the evolution of the traffic environment in simulation always decreases the applications' performance, such as generating the unexpected stop-and-go wave, and the generation error will grow both temporally and spatially.

Hence, we define a new mission named "platoon trajectories generation" to reproduce realistic full trajectories. Specifically, platoon trajectories generation is a generalized car-following process in a platoon, which aims to generate the actual full trajectories of the platoon of HDVs as realistic as possible. Though many researches have been conducted in trajectory prediction [17], [22], [23], the trajectory generation has great differences in the aspect of information utilization and availability. To systematically describe the differences, we use Fig. 1 as an example to illustrate. Specifically, as shown in Fig. 1(a), during the studied period (from past to now), given the trajectory of the leading vehicle of the platoon (green curve of vehicle 1) and followers' initial position (green cross dots of vehicle 2~7), platoon trajectories generation mission will generate the trajectories of all following vehicles (vehicle 2-7, red dashed curves) in the platoon, which is commonly used to represent the actual trajectories under limited observation. Different from the platoon trajectories prediction mission (Fig. 1(b)), which predicts the future trajectories according to fully observed data (from past to now), generation mission utilizes partial historical data to estimate unobserved trajectories on or before the present time. In the micro-traffic simulation platform, every simulation step is an extreme case of platoon trajectories generation, where the length of the studied period equals the simulation interval and the trajectory of the leading vehicle has degenerated to the initial position. In other words, the research on the platoon trajectories generation mission helps in analyzing the long-term trajectory accuracy of vehicles in the platoon in traffic simulation.

However, directly applying traditional methods (e.g. using IDM/LSTM-based model to decide vehicle's behavior at the next simulation step) to the platoon trajectories generation mission will induce a fatal accuracy problem in traffic simulation platform, the "error propagation" problem. Specifically, error propagation refers to a phenomenon that the error between the generated and actual trajectories accumulates and propagates both in temporal and spatial dimensions, and the error propagation effect will be more significant during traffic oscillation. Fig. 1(a) illustrated the actual trajectories (green solid curves) and generated trajectories (red dashed curves) during a traffic oscillation via a numerical generation method by a typical LSTM-based car-following model [11]. While the trajectory of the first following vehicle (vehicle 2) is well estimated, the error of following trajectories is gradually amplified vehicle by vehicle. The existing neural network structures of the car-following model largely ignore the error propagation problem without considering vehicles' interconnection in the aspect of the platoon.

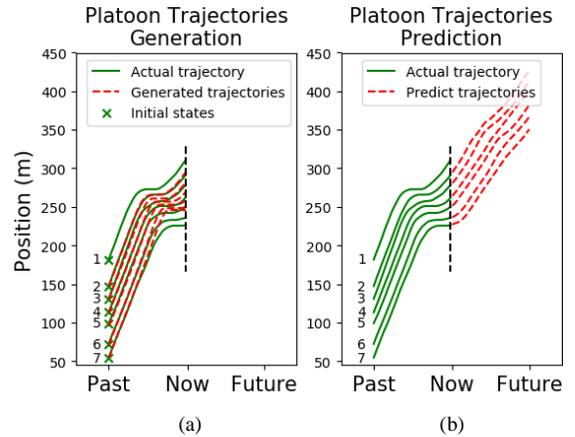

Fig. 1. Platoon trajectories generation and trajectories prediction

In order to solve the accuracy issue, this paper analyzes and summarizes the fundamental causes of the error propagation problem in both temporal and spatial dimensions. Basing on the analysis, we propose an LSTM-based interconnected car-following model (Int-LSTM) to solve the error propagation problem. The result illustrates that the proposed model drastically reduces the error of generated trajectories. The contribution of this paper is threefold: (1) propose the platoon trajectories generation mission and analyze the main cause of error propagation problem; (2) devise a sampling mechanism to overcome the temporal error propagation problem in the training process of the car-following model; (3) propose an interconnected LSTM structure and use platoon-level training strategy to reduce the spatial error propagation problem.

The rest of the paper is organized as follows: Section II analyses and summarizes the fundamental causes of the error propagation problem using the block diagram representation of the car-following model. Section III elaborates on the unidirectional interconnected structure of the car-following models and discusses the training method of the model. Section IV adopts empirical trajectories data to evaluate the performance of the proposed method, and analyses the results. The last section concludes the research and points out the development direction.

## II. CAR-FOLLOWING MODELS BLOCK DIAGRAM REPRESENTATION AND ERROR PROPAGATION ANALYSIS

During the past decades, various car-following models [4], [15]–[19] have been proposed and improved in order to accurately represent the principle of driving behavior. In general, most researches decompose the car-following model into several successive decision steps. To be specific, at the $t$th decision step, it decides how to maneuver the vehicle in a short time interval $\Delta t$ before the next decision step. To better illustrate the process, we will firstly abstract the general decision step of the car-following model to a block diagram. As shown in Fig. 2, the block with few input/output arrows represents the decision step of a vehicle at a specific time step. At each decision step, the driver will make a decision according to the following inputs:

- *The immediate state of the vehicle and its leading vehicle, $s_F(t)$ and $s_L(t)$*: Vehicle's state at the $t$ th decision step $s(t)$ is a vector, which consists of the position $x(t)$, velocity $v(t)$, and acceleration $a(t)$. The subscript $F$ and $L$ is the abbreviation of the "Following" and "Leading" vehicle. For the $i$ th vehicle in the platoon, $s_i(t) = (x_i(t), v_i(t), a_i(t))$, $s_F(t) = s_i(t)$, and $s_L(t) = s_{i-1}(t)$.
- *The historical information $h(t-1)$*: Due to the response delay of drivers, historical information is also considered in some car-following models. It may represent a specific vehicle state, such as the historical position in Newell's car-following model [24], or calculated hidden state in Neural Networks (NN) based car-following model [17].

The output of the block is the state $s_F(t+1)$, which will be reached in the next decision step, and the updated historical information $h(t)$ for future decisions. For the sake of convenience, the current state $s_F(t)$ is also included in the output. Inside the block, the general car-following model could be formulated as follows:

$$a_F(t+1) = f(s_F(t), s_L(t), h(t-1)) \quad (1\text{-}a)$$
$$v_F(t+1) = v_F(t) + a_F(t+1)\Delta t \quad (1\text{-}b)$$
$$x_F(t+1) = x_F(t) + v_F(t+1)\Delta t \quad (1\text{-}c)$$
$$h(t) = g(s_F(t), s_L(t), h(t-1)) \quad (1\text{-}d)$$

where (1-a) represents the decision function with inputs $s_F(t)$, $s_L(t)$, and $h(t-1)$; (1-b) and (1-c) indicates the uniform acceleration and velocity process; (1-d) denotes the update function (e.g. position shift in Newell's car-following model; the update function of the hidden state in NN-based model) of the historical information. The output of the block $s_F(t+1)$ will be calculated through those three equations.

Based on block diagram representation, we will further discuss and summarize the reason of the error propagation problem among current trajectory simulations, which mainly caused by the training method and model structure. The error propagation problem is decomposed in the temporal and spatial dimensions. In the temporal dimension, the training method of the car-following model used in previous researches brings the mismatch between the training and inference process. In the spatial dimension, the structure of traditional models ignores the topology in the platoon.

Training and inference are two critical parts of the platoon trajectories generation. Specifically, training refers to calibrating the parameters of a car-following model by feeding with empirical driving data. Correspondingly, inference denotes using the trained model to generate trajectories of the following vehicles. Fig. 3 demonstrates the typical training process, which could be demonstrated by an array of car-following blocks. "Vehicle $i$" indicates the $i$th vehicle in the platoon. The corresponding row of the array indicates the iterative decision-making process of the vehicle in the temporal dimension. Each column of array denotes the sequential car-following process in spatial dimension at a specific decision step. In the training process, the model uses the actual state of both the leading vehicle (e.g. $s_1(1)$) and following vehicle (e.g. $s_2(1)$) as input, and updates the model's parameters according to the difference between the actual state (e.g. $s_2(2)$) and the generated state (e.g. $\hat{s}_2(2)$). However, during the inference process, the actual state (e.g. $s_2(2)$) is unavailable, and are thus replaced by the generated state (e.g. $\hat{s}_2(2)$). Due to the bias between generated and actual state, the error is added into the inference process. Both temporal and spatial dimension will suffer from the error propagation problem.

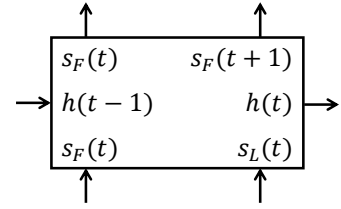

Fig. 2 General decision step of the car-following model

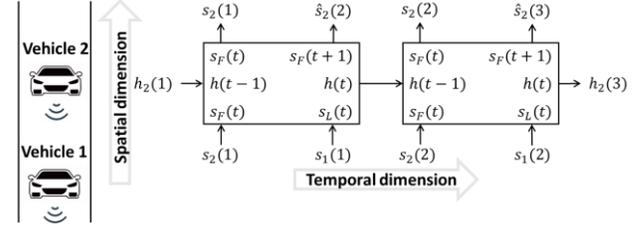

Fig. 3 Training process of the car-following model

In order to better understand the error propagation problem, we will provide a few examples. For simplicity's sake, we adopt a typical LSTM-based car-following model mentioned in [11] as an illustration, which feeds $v_i(t)$, $\Delta v_i(t)$, and $\Delta x_i(t)$ as inputs into LSTM and outputs the velocity $\hat{v}_i(t+1)$. The generated position of the next decision step $\hat{x}_i(t+1)$ is calculated by (1-c). The model training algorithm used in the LSTM-based car-following model is shown in Algorithm 1. At the $k$ th training epoch, we firstly calculate the generated trajectory $\hat{x}_2^{(n)}(t|\theta)$ using the LSTM-based car-following model $f_{LSTM}(\cdot)$ (line 3-8 in Algorithm 1). Then, the parameter $\theta_k$ of LSTM-based model is updated by minimizing loss function of Mean Squared Error (MSE) of position difference measure (line 9 in Algorithm 1), where $T$ denotes the total number of decision steps; $N$ indicates the number of vehicle pairs in the dataset; $p(\cdot)$ denotes the gradient function used in various gradient method in optimization, such as SGD and Adam [25]; $\alpha$ is the learning rate.

We will first demonstrate how traditional training methods cause the temporal error propagation problem by comparing the lose ($L(\theta)$ in Algorithm 1) in the training and inference process. In general, along with the number of training epochs increasing, both the training and inference error should decrease gradually and finally converged to a stable value. However, Fig. 4 demonstrates a totally different result, where the inference error has no significant drop and even shows a slight growth trend at the end of the training. Besides, the minimum training error (0.74) is obviously smaller than the minimum inference error (6.7). All of those phenomena illustrate that the error is propagated and amplified in the inference process. We believe that the training error is isolated in the training process because of using the actual state as the input of the car-following model. However, in the inference process, the error from previous steps will be propagated to the next decision step in the temporal dimension.

In the spatial dimension, vehicles' behavior is largely depended on their leading vehicles. The accuracy of the car-following model will also dominate the performance of platoon trajectories generation. If the trained model performs an aggressive driving behavior [8], the following vehicle will not slow down until the gap between the following and leading vehicle becomes too close. Similarly, a timid driving behavior will lead to a huge following gap. An illustrative example of the spatial error propagation problem is shown in Fig. 5. We use the trained LSTM-based car-following model to generate the trajectory of a platoon. The heat map of the inference error in Fig. 5 illustrates the phenomena of error accumulation and propagation problems toward upstream. Each row represents the inference error of a specific vehicle. Though the error of the first following vehicle (vehicle 2) is negligible during the studied period, the inference error begins to be propagated along with the spatial dimension and becomes worsened at the right bottom corner of the heat map.

In summary, the main sources of the error propagation problem come from two aspects. First, the traditional training method leads to temporal error propagation in the inference process because of ignoring the bias in the generated states. Second, the structure of the traditional car-following model isolates the spatial topology in the platoon and therefore causes spatial error propagation.

### III. UNIDIRECTIONAL INTERCONNECTED LSTM-BASED CAR-FOLLOWING MODEL

Based on the analysis in the previous section, we propose a unidirectional interconnected LSTM-based car-following model (Int-LSTM) to solve the error propagation problem. The Int-LSTM model has the following major improvements on the basis of LSTM-based car-following model: (1) the model involves scheduled sampling (SS) mechanism to bridge the gap between the training and inference process, and decrease the temporal error propagation problem; (2) enriching the LSTM-based model with topology feature to overcome the spatial error propagation problem, since the vehicle platoon is a unidirectional interconnected system [2], and utilizing platoon-level training strategy to fuse both dimensional improvements. The interconnected structure of the Int-LSTM model is shown in Fig. 6, where 'SS' indicates the schedule sampling module. More details will be introduced in this section.

The scheduled sampling mechanism mainly solves the

**Algorithm 1** Model training LSTM-based car-following model

1. Input: Vehicle pairs data, network parameter $\theta_k$
2. Output: updated $\theta_{k+1}$
3. For each vehicle pair $n$:
4.   Initialize $h(0) = 0$
5.   For each decision step $t$:
6.     $a_2(t+1) = f_{LSTM}(s_2(t), s_1(t), h(t-1)|\theta_k)$
7.     $v_2(t+1) = v_2(t) + a_2(t+1)\Delta t$
8.     $x_2(t+1) = x_2(t) + v_2(t+1)\Delta t$
9. Update $\theta_{k+1} \leftarrow \theta_k - \alpha \cdot p(L(\theta_k))$, where

$$L(\theta) = \frac{1}{N}\sum_{n=1}^{N}\frac{1}{T}\sum_{t=2}^{T}[x_2^{(n)}(t) - \hat{x}_2^{(n)}(t|\theta)]^2$$

10. Return $\theta_{k+1}$

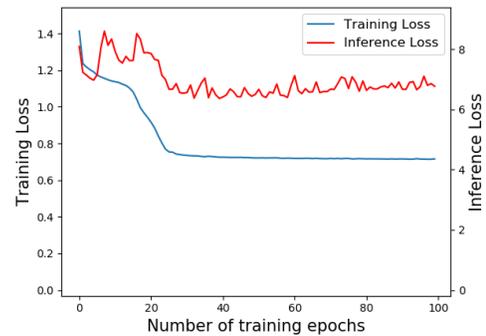

Fig. 4 Temporal error propagation

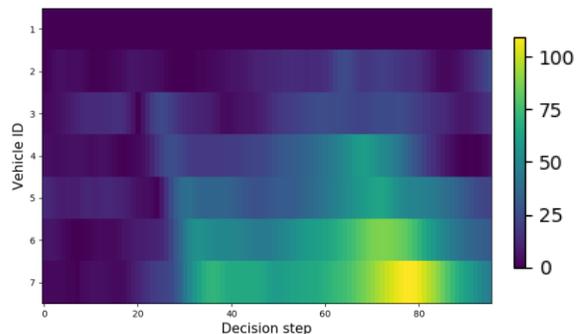

Fig. 5 Heat map of inference error in spatial dimension

temporal error propagation problem. During the training process, it will randomly decide whether we use actual or generated state [26]. Scheduled sampling ensures that the car-following model could be fitted by empirical data (actual state) while involving generated state to fine-tune itself. As shown in Fig. 6, the input of the upper right car-following block is decided by two scheduled sampling modules. The module's input with a solid arrow indicates the actual state (e.g. $s_3(2)$), others are generated state (e.g. $\hat{s}_3(2)$). At the $k$th training epoch, the scheduled sampling module will output the actual state with probability $\epsilon_k$, or generated state with probability $1 - \epsilon_k$. By scheduling the $\epsilon_k$, we can modify the probability of feeding generated state into the model during the training process. Intuitively, the network is not well trained at the beginning of training. Sampling from the generated state will lead to slow convergence. On the contrary, at the end of the training, the generated state should be feed into the network in order to reduce the error propagation problem. Therefore, as training progress, the probability of sampling from the generated state should increase, in other words, we need to use a schedule to decrease $\epsilon_k$ gradually. In this paper, decrease schedules could be expressed by,

$$\epsilon_k = \begin{cases} med(0, decay(k), 1), & k \in [0, epoch] \\ 0, & others \end{cases}. \quad (2)$$

where $epoch$ determine when the probability decrease to 0; the decay schedule function $decay(k)$ is related to the number of training epochs $k$. TABLE I lists three typical decrease schedules functions: linear decay, exponential decay, and inverse sigmoid decay [26]. $w$ and $c$ represent the decrease rate and offset of the function, respectively. Except for those three functions, Fig. 6 also visualizes the situation of constant $\epsilon_k$. When $\epsilon_k = 1$ for all training epochs, the input of the car-following model will always from the actual state, the scheduled training process becomes the traditional training process in Fig. 3. Similarly, when $\epsilon_k = 0$, all generated states will be feed to the model.

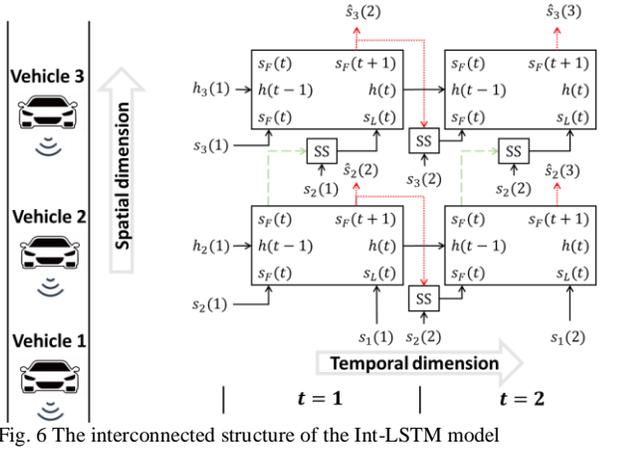

Fig. 6 The interconnected structure of the Int-LSTM model

TABLE I
TYPICAL DECAY SCHEDULE FUNCTIONS

| Decay schedule | $f(n)$ |
| --- | --- |
| Linear | $wn+c$ |
| *Exponential* | $w^n + c$ |
| *Inverse sigmoid* | $1 - \dfrac{1}{1+e^{-w(n-c)}}$ |

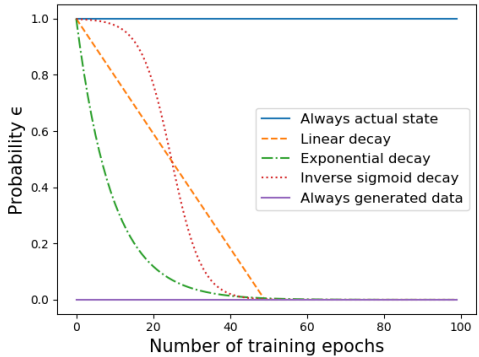

Fig. 7 Example of decay schedules

Due to the accumulation effect of the platoon trajectories

**Algorithm 2** Parameter updating in the proposed car-following model

1. Input: Platoon vehicle data, network parameter $\theta_k$
2. Output: updated $\theta_{k+1}$
3. For each platoon $n$:
4.   For each following vehicle $i$:
5.     Initialize $h(0) = 0$
6.     Calculate $\epsilon_k = decay(k)$
7.     Generate random binary arrays $b_1, \ldots, b_T$ using $\epsilon_k$
8.     For each decision step $t$:
9.       $a_i(t+1) = b_t \cdot f_{LSTM}(s_i(t), s_{i-1}(t), h(t-1)|\theta_k) + (1-b_t) \cdot f_{LSTM}(\hat{s}_i(t), \hat{s}_{i-1}(t), h(t-1)|\theta_k)$
10.      $v_i(t+1) = v_i(t) + a_i(t+1)\Delta t$
11.      $x_i(t+1) = x_i(t) + v_i(t+1)\Delta t$
12. Update $\theta_{k+1} \leftarrow \theta_k - \alpha \cdot p(L(\theta_k))$, where
$$L(\theta) = \frac{1}{N}\sum_{n=1}^{N}\frac{1}{T}\sum_{t=2}^{T}\frac{1}{I}\sum_{i=2}^{I}\left[x_i^{(n)}(t) - \hat{x}_i^{(n)}(t|\theta)\right]^2$$
11. Return $\theta_{k+1}$

generation, training car-following model under the vehicle-level scenario (only one following vehicle is considered, as shown in Fig. 3) will not help to eliminate the spatial error. In the spatial dimension, interconnected structure assembles all vehicles' car-following models (extending more vehicles in spatial dimension in Fig. 6) and ensures that the car-following model could be trained in platoon-level. With the help of interconnected structure, once the temporal/spatial error propagation (along with the direction of green/red dashed arrows) occurs during the training process, it will be weakened immediately by modifying the parameter of the car-following model. It is noted that both the training and inference process could use the same structure, but the decay functions in the scheduled sampling module are different. In the inference process, we have to set $\epsilon_n = 0$, due to the lack of the actual state. However, various decay functions could be used for better training results in the training process.

Different from the training method (Algorithm 1) for the LSTM-based car-following model, Int-LSTM adopts Algorithm 2 to train the model parameters. In Algorithm 2, line 6-9 reflect the selection process of the scheduled sampling module. The algorithm generates random binary arrays $[b_1, \ldots, b_T]$ to decide using either actual or generated state in training. When updating the model parameters, the loss function in Algorithm 2 adopts platoon-level MSE of position difference measure (line 12 in Algorithm 2), where $I$ indicates the number of vehicles in the platoon.

In summary, the difference between the LSTM and Int-LSTM model is reflected in the training process. LSTM isolates the error generated and can only learn the driving behavior from vehicle pairs (platoon length is 2). Int-LSTM breaks the constraint of that by combining scheduled sampling and platoon-level training with LSTM. As a result, both temporal and spatial error propagation will be decreased in the training process.

IV. EXPERIMENTS

In this section, we will conduct several experiments to evaluate the performance of the Int-LSTM car-following model. We will firstly compare different decay functions of the scheduled sampling module so as to determine the best function for the training of the car-following model. Then, the performance of the existing and proposed car-following model will be compared. Furthermore, we will also discuss the contribution of both scheduled sampling and platoon-level training for reducing error propagation.

*A. Experiment Setting*

Given the trajectory of the first leading vehicle (vehicle 1) and the initial state (e.g. position and speed) of other following vehicles, the tested methods will generate the trajectory of the whole platoon. Considering both the amount of data and validity of the experiment, we design a car-following scenario of a 5-vehicle platoon for both model training and evaluation in our experiments, where the lane changing behavior is not included. The studied period of car-following lasts 20s; the time interval between two decision steps is 0.5 seconds. As recommended by [5], the Int-LSTM model has three hidden layers that contain 10, 10 and 5 neurons, respectively. The inputs and outputs are the same as the LSTM in section II.

The experiments metrics used in this paper include:
- Mean Absolute Error (MAE),
- Mean Maximum Absolute Error (MMaAE),
- Cumulative distribution of Absolute Error (AE),
- Cumulative distribution of Platoon Maximum Absolute Error (PMaAE).

MAE of the position difference indicates the inference error of the whole platoon during a specific period; MMaAE of the position difference is designed to represent the level of the error propagation; cumulative distribution of AE and PMaAE reflects more detailed error distribution. The followings are the expressions of them,

$$MAE = \frac{1}{N}\sum_{n=1}^{N}\frac{1}{T}\sum_{t=2}^{T}\frac{1}{I}\sum_{i=2}^{I}|x_i^{(n)}(t) - \hat{x}_i^{(n)}(t)| \quad (3)$$

$$MMaAE = \frac{1}{N}\sum_{n=1}^{N}\max_{i,t}|x_i^{(n)}(t) - \hat{x}_i^{(n)}(t)| \quad (4)$$

$$AE = |x_i^{(n)}(t) - \hat{x}_i^{(n)}(t)|, \quad (5)$$

$$PMaAE = \max_{i,t}|x_i^{(n)}(t) - \hat{x}_i^{(n)}(t)| \quad (6)$$

where $x_i^{(n)}(t)$ and $\hat{x}_i^{(n)}(t)$ represents the actual/generated position of the $i$th vehicle at the $t$th time step in the platoon $n$.

The empirical trajectory data used in this paper comes from the Next Generation Simulation (NGSIM) [27]. Due to the noise in the dataset, we adopt the reconstructed NGSIM dataset from Montanino et al. [28]. The reconstructed data includes vehicles' trajectories for I-80, California from 4:00 pm to 4:15 p.m. on April 13, 2005. To avoid the interruption from the frequent lane-changing operation, trajectories on lane 1~4 are considered. Besides, we will select trajectories from the specific platoon, which consists of at least 5 vehicles and persists at least 20s, to reserve more car-following features. In training, in order to cover the driving behavior under various traffic scenarios, data from lane 1 (HOV, free flow) and lane 3~4 (congested flow) will be used (3752 platoons). Finally, we will use the data from lane 2 (congested flow) for evaluation (73 platoons).

TABLE II
METRICS OF VARIOUS DECAY FUNCTIONS

| Decay Function | MAE | MMaAE |
| --- | --- | --- |
| Always actual state | 15.7 | 53.1 |
| Linear decay | 14.2 | 47.0 |
| Exponential decay | 13.7 | 46.5 |
| Inverse sigmoid decay | **13.5** | 46.3 |
| Always generated state | 14.2 | **46.0** |

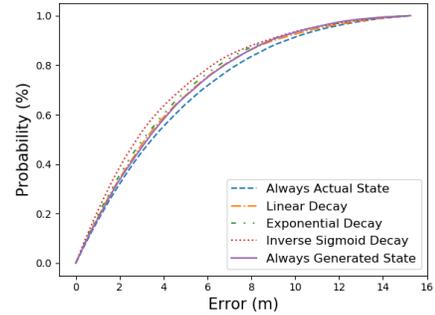

(a) Absolute error

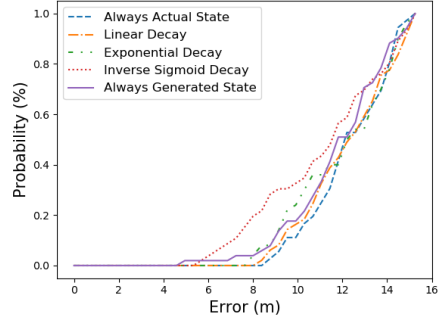

(b) Platoon maximum absolute error

Fig. 8 Error distribution under various decay functions

*B. Selection of decay function*

To find the best decay function, we will test the Int-LSTM model with 5 typical decay functions mentioned in section III. The number of training epochs is set as $epoch = 100$. The following lists the parameters of different decay functions:
- Always actual state: $\epsilon_i = 1$,
- Linear decay: $k = -2/epoch$, $c = 1$,
- Exponential decay: $k = 0.9$, $c = 0$,
- Inverse sigmoid decay: $k = 1/4$, $c = epoch/4$,
- Always generated state: $\epsilon_i = 0$.

As we have known, always sampling from actual or generated are two extreme cases in scheduled sampling. The former is equivalent to the training process of LSTM without scheduled sampling, whereas the latter directly uses the inference way for training.

TABLE II shows the results of MAE and MMaAE on different decay functions. Int-LSTM with inverse sigmoid decay tends to obtain the best performance, but does not obviously superior to other decay functions. Compared with constant $\epsilon_i$, decreasing decay function (e.g. linear, exponential, and inverse sigmoid decay) successfully trained a model more resilient to failures. Without the help of scheduled sampling, always sampling from actual state has the worst performance on all metrics because of ignoring the temporal error propagation problem. On the contrary, though always sampling from the generated state does not achieve the minimum MAE, it reduces the MMaAE from 53.1 to 46.0, which also proves the importance of bridging the gap between training and inference.

Fig. 8 compares the cumulative distribution of AE and PMaAE under various decay functions. The area under the cumulative distribution curve reflects the model's performance. Bigger area means the model would have a higher probability to generate trajectory with small error. Therefore, inverse sigmoid decay performs better under both metrics. Though the area under the AE curve (dotted curve in Fig. 8(a)) is not obviously enhanced under inverse sigmoid decay, the PMaAE is reduced significantly (dotted curve in Fig. 8(b)). As we have known, error propagation always leads to bigger maximum error. Less PMaAE means that inverse sigmoid decay could effectively solve the error propagation problem. According to the result mentioned above, in the rest of this paper, we will use the inverse sigmoid decay function in the training process of the Int-LSTM model.

*C. Performance comparison*

In order to demonstrate the improvement of the Int-LSTM model, we also conduct several car-following models as baselines:
- IDM,
- LSTM-based model,
- LSTM-based model with scheduled sampling,
- LSTM-based model with platoon-level training.

The parameters of the IDM are listed in TABLE III, which is suggested in [11]. The LSTM-based model has the same neurons as Int-LSTM, but without scheduled sampling and platoon-level training.

A comprehensive result is shown in TABLE IV, where Int-LSTM has less inference error in both MAE/MMaAE. Compared with IDM/LSTM-based model, the MAE/MMaAE of Int-LSTM is reduced at least by 20%. Fig. 9 compares the generated trajectories during a traffic oscillation by different car-following models. In the result of the IDM/LSTM-based model, while the trajectory of the first following vehicle is well estimated using both models, the error of other following trajectories is gradually amplified vehicle by vehicle. Especially, the LSTM-based model generates a more serious error propagation problem, which is believed to be caused by overfitting problem in training. A similar result could be observed in Fig. 10, which illustrates the error distribution of various models. LSTM-based model prefers to have a bigger error, compared with other models. Especially in PMaAE (Fig. 10 (b)), the maximum error of the generated trajectories by the LSTM-based model is tending to be bigger.

Whereas the traditional LSTM-based model performs worse than IDM in all metrics, the inference error has been improved to the same level as IDM (from 25.1 to 17.1) after combining the LSTM-based model with platoon-level training. To verify which improvement benefits more in reducing the error propagation problem. We combine the LSTM-based model with scheduled sampling and platoon-level training respectively. The result in TABLE IV shows that, with the help of scheduled sampling, the LSTM-based model is improved in all metrics. By contrast, the platoon-level training does not provide such a great improvement. Fig. 11 compares the generated platoon trajectories of different LSTM variations under the traffic oscillation scenario. In Fig. 11(a) (LSTM-based model), the error propagation problem occurs from the 20s to 50s. Even with the help of the platoon-level training (shown in Fig. 11(b)), such a problem is not been effectively solved. However, involving scheduled sampling makes the trained car-following model present a better behavior (in Fig. 11(c)), whereas the last vehicle in the platoon moves behind the actual trajectory at the end of the studied period. Finally, the Int-LSTM model merges both improvements and generates the best platoon trajectories (in Fig. 11(d)). Just like the analysis in section II, the spatial error propagation problem comes from the accumulation of the error in the temporal dimension. Once the temporal error is eliminated by scheduled sampling, the error is less likely to be propagated in the spatial dimension. Therefore, scheduled sampling benefits more in reducing error propagation problem than platoon-level training. In summary, scheduled sampling plays an important role in reducing the temporal error propagation problem; the platoon-level training is used to fine-tune trajectories from the perspective of the platoon. The combination of both improvements will make Int-LSTM have a more robust performance.

TABLE III
PARAMETERS OF IDM

| Parameter | Value | Units | Description |
|---|---|---|---|
| $a$ | 1.4 | $m^2/s$ | Maximum acceleration |
| $b$ | 2.0 | $m^2/s$ | Desired deceleration |
| $V_0$ | 30 | $m/s$ | Desired velocity |
| $g_{jam}$ | 2 | $m$ | Minimum gap in the jam |
| $T$ | 1.5 | $s$ | Safe time headway |

TABLE IV
PERFORMANCE OF DIFFERENT MODELS

| Models | MAE | MMaAE |
|---|---|---|
| IDM | 17.2 | 58.5 |
| LSTM | 25.1 | 76.5 |
| LSTM+Platoon-level training | 17.1 | 56.9 |
| LSTM+Scheduled sampling | 15.4 | 51.9 |
| Int-LSTM | **13.5** | **46.3** |

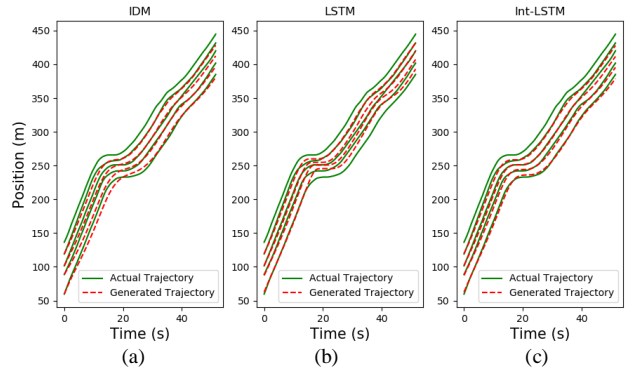

Fig. 9. Generated trajectory comparison among different methods

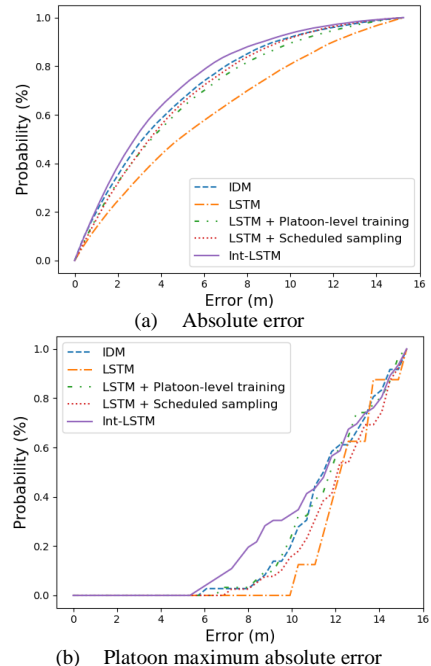

(a) Absolute error

(b) Platoon maximum absolute error

Fig. 10 Error distribution of different models

## V. Conclusion

In this paper, we proposed a platoon trajectories generation mission to relieve the error propagation phenomenon prevailing in microsimulation software. Firstly, we analyzed the main cause of the error propagation problem in the platoon trajectories generation mission based on the car-following model block diagram analysis. Based on that, we proposed a unidirectional interconnected LSTM-based car-following model to generate platoon level full trajectories. The model adopts two improvements on the basis of the LSTM-based car-following model. We utilize the scheduled sampling technique to solve the error propagation in the temporal dimension. Both actual and generated state will be used for model fitting during the training process. It not only makes the trained model learn the actual behavior but also fixes the mismatch between training and inference. Then, platoon-level training is adopted in the training process. It helps in extracting platoon-level features and efficiently reducing the spatial error propagation problem. The empirical experiments show that the proposed model significantly reduce the error propagation problem in both the temporal and spatial dimension. Compared with the traditional LSTM-based car-following model, the Int-LSTM model has almost 40% less inference error. Besides, the inverse sigmoid decay function is most appropriate for the training of the car-following model.

Int-LSTM model outperforms model-based car-following models in the platoon trajectories generation mission. Because of that, it provides an opportunity for the NN-based car-following model to be applied in related transportation applications. The high-accuracy simulation of HDV's behavior will also benefit the research of both microscopic traffic modeling. In the future, more researches could be developed on the basis of this paper. Firstly, the combination of traffic flow theory and neural network might become a new research hotspot. It would be possible to explain complicated traffic phenomena by the NN-based car-following model by adding attention layer [29], [30]. Secondly, the safety issue of the NN-based car-following model is still unsolved. It is difficult to teach the neural network the physical rule (e.g. no collision) from the empirical data. Then, typical driving scenarios, such as lane changing and signal intersection, should also be considered in order to enrich the comprehensive driving behavior model. Finally, the findings in this paper give us an opportunity to build a more realistic mixed platoon simulation platform, which will be used to evaluate and optimize the performance of autonomous vehicles' controlling strategy in the simulated mixed environment.

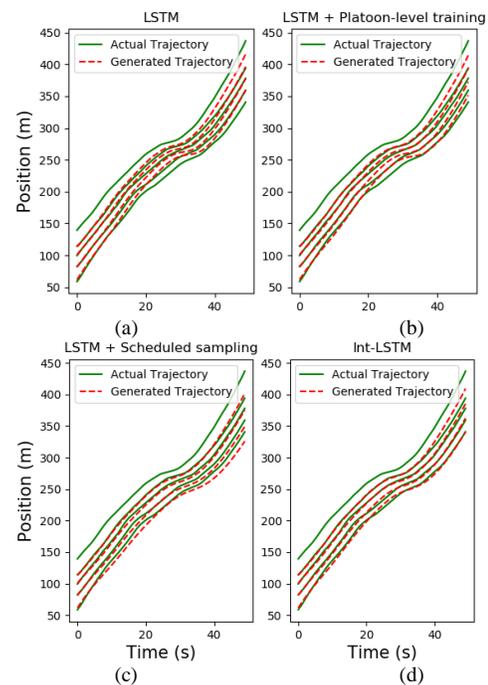

Fig. 11 Trajectory of different LSTM variations